\documentclass[3p,times,procedia]{elsarticle}

\usepackage{amsmath}
\usepackage{algorithmic}
\usepackage{algorithm}
\usepackage{array}
\usepackage[caption=false,font=normalsize,labelfont=sf,textfont=sf]{subfig}
\usepackage{textcomp}
\usepackage{stfloats}
\usepackage{url}
\usepackage{verbatim}
\usepackage{graphicx}
\usepackage{hhline}
\usepackage{multirow}
\usepackage{natbib}

\usepackage{amssymb}

\usepackage[figuresright]{rotating}

\begin{document}

\begin{frontmatter}




\title{LocalEyenet: Deep Attention framework for Localization of Eyes}


\author{Somsukla Maiti, Akshansh Gupta}

\address{CSIR-Central Electronics Engineering Research Institute, Pilani, Rajasthan, India-333031\\
somsukla@ceeri.res.in, somsuklamaiti@gmail.com; akshanshgupta@ceeri.res.in}

\begin{abstract}
Development of human machine interface has become a necessity for modern day machines to catalyze more autonomy and more efficiency. Gaze driven human intervention is an effective and convenient option for creating an interface to alleviate human errors. Facial landmark detection is very crucial for designing a robust gaze detection system.  Regression based  methods capacitate good spatial localization of the landmarks corresponding to different parts of the faces. But there are still scope of improvements which have been addressed by incorporating attention. \\
In this paper, we have proposed a deep coarse-to-fine architecture called LocalEyenet for localization of only the eye regions that can be trained end-to-end. The model architecture, build on stacked hourglass backbone, learns the self-attention in feature maps  which aids in preserving global as well as local spatial dependencies in face image. We have incorporated deep layer aggregation in each hourglass to minimize the loss of attention over the depth of architecture. Our model shows good generalization ability in cross-dataset evaluation and in real-time localization of eyes.
\end{abstract}


\end{frontmatter}


\section{Introduction}
Modern machines are user-friendly and keep the possibility of human machine interaction open. Human interaction provides an effective way of controlling the machine with ease and reduces the risk of error. Gaze driven human machine interfaces have become essential for smooth controlling of machine parts with no physical intervention. The need of gaze control has been addressed since the past two decades in assistive robotic systems \cite{shafti2019gaze} for disable persons and in controlling the robotic arms in robotic surgical system \cite{ali2008eye}. Tracking the eye gaze requires accurate localization of the facial landmarks. Face landmark detection has been playing a significant role in current human machine interface applications, such as gaze tracking for autonomous vehicles \cite{jessee2017gaze}, face tracking \cite{wang2021visual}\cite{digiacomo2020head}\cite{liu2021development} and facial expression analysis \cite{cordea2008three}\cite{mohan2020facial}.

Localization of landmark points on the face helps us in performing face alignment in an effective manner. While designing a robust Gaze controlled interface, it is mandatory to perform precise localization of eyes with high accuracy and low latency. Simultaneously, landmark detection methods are required to handle different real-time challenges, such as low lighting condition, face occlusion and fast movement of head that lead to variation in pose.
Since past two decades, with the evolution of new methods in deep learning, there has been a significant improvement in developing more robust solutions. Convolutional neural networks (CNNs) have been widely used for facial landmark detection for different applications. Availability of large annotated face databases in different environmental conditions has made the task easier. Even with small dataset, different data augmentation techniques have been proved really efficient in providing better generalization to the solution. 
Coarse to fine regression type of architectures has shown the most prominent localization performance. These techniques learns the coarse features over a shape space at the shallow layers and subsequently learns the fine features at deeper layers \cite{zhu2015face} by cascading several deep CNN \cite{sun2013deep}. Cascaded hourglass and Unet architectures have been able to provide good generalization but still have some issues. The heatmap generated using most of these state of the art architectures do not compute the local attention after each layer. This affects the learning of correlation between features from the shallow layers and the deep layers. \\
We here aim to provide a solution that solves the following issues by designing different modules.
\begin{itemize}
	\item We have defined a self-attention module between the individual hourglasss modules, which learns the local spatial attention and the global attention to estimate the precise location of the landmarks.  
	\item We have designed a deep layer aggregation module to learn the feature dependencies over the depth of architecture while parsing the features across network.
	\item We have used differentiable Soft-argmax \cite{luvizon20182d} \cite{luvizon2019human} to make the framework an end-to-end trainable architecture to determine the coordinates $(x,y)$ of each facial landmark from the attention heatmaps.
\end{itemize}

\section{Related Work}
Facial landmark detection and localization has been an active area of research for aligning faces. There have been three broad working techniques \cite{wu2019facial}, a. holistic techniques, b. Constrained Local Model (CLM) based techniques  and c. regression-based techniques. Holistic models \cite{tzimiropoulos2015project} generate appearance model of the face images by first building a shape model and then mapping the features in the shape model. Active appearance models \cite{edwards1998interpreting} \cite{cootes2001active} have been one of the widely used holistic approach that computes and updates the shape and appearance coefficients and the parameters for affine transformations for detection of the facial landmarks. 
CLM based techniques generate global shape model and local appearance models for each landmark points \cite{xiong2013supervised}. Local appearance models help in minimizing the error due to variation in illumination and occlusion. Baltrusaitis et. al presented a Constrained Local Neural Field (CLNF) \cite{baltrusaitis2013constrained} that take care of the feature detection problem in wild scenario with less illumination and blurred faces. Recently Weighted Iterative Closest Point (ICP) based surface matching \cite{su2022facial} and hierarchical filtering strategy \cite{jin2020towards} have been used to reduce the effect of noise in face registration. Researchers have recently proposed a semi-supervised method \cite{wan2021robust}, called self-calibrated pose attention network (SCPAN), that computes Boundary-Aware
Landmark Intensity (BALI) fields corresponding to a boundary and the landmarks closest to the boundary. They have also extended their work by proposing an implicit multiorder correlating geometry-aware (IMCG) model \cite{wan2021robust1} \cite{wan2021robust2}, that uses spatial and channel correlations to attain the local as well as global features.

On the other hand, the regression based techniques mainly focuses on learning a model that maps the facial landmarks on the images, in spite of developing global and local face models.   The regression models can be mainly classified into two categories, viz. i. Cascaded regression techniques and ii. Coarse-to-fine techniques.

\subsection{Cascaded regression techniques}
Cascaded regression methods have been proved to be very efficient and it mainly relies on extracting features at each stage and optimizing the regression parameters for estimating the positions of facial landmarks. Most of the cascaded regression methods relies on hand-crafted features. Local binary features (LBF) \cite{cao2014face} and Histogram of Oriented Gaussians (HoGs) have been the most used hand-crafted features that are used for the facial landmarks. 
In \cite{ren2014face}, LBFs have been extracted using random forests and linear regression model is learned  for each landmark position. However development of regression models based on dedicated feature extraction methods tend to make the solution more sensitive to the given data. Thus most of the time, it fails to provide a generalized solution in scenarios with varied illumination conditions, occlusions and varied poses. The current cascaded regression models aim to develop a deep cascaded end-to-end architecture to determine the landmark coordinates \cite{burgos2013robust} \cite{zhu2019robust} \cite{feng2018wing}. In \cite{li2022towards}, the authors have developed a Deformable Transformer Landmark Detector (DTLD) model that preserves the local spatial structure of the face images and improves the localization accuracy. Lai et al.\cite{lai2018deep} have included recurrent neural network (RNN) modules to learn the dependency of the features generated by the cascaded CNN modules. Weng et al have developed a cascaded deep autoencoder network (CDAN) \cite{weng2016learning}, that learns the feature representation at the global and the local stages simultaneously in cascaded manner. 

\subsection{Coarse-to-fine techniques}
Coarse-to-fine techniques have been immensely used by developing different deep learning architectures \cite{zhu2015face} \cite{li2020structured} \cite{sun2019deep} \cite{zhang2014coarse} \cite{trigeorgis2016mnemonic} for generating accurate prediction of the landmark points. Dapogny et al. has defined spatial softargmax \cite{dapogny2019decafa} to generate landmark-specific attention. Kim et al developed an extended version of MTCNN model, called EMTCNN model \cite{kim2020augmented} where they have dilated convolution and CoordConv to improve the localization accuracy.  Several works has been performed, where facial recognition, head pose determination and other activities has been performed along with facial landmark detection. In such applications, shared CNN \cite{zhang2014facial} have been developed that uses same set of  feature maps and enables better representation learning. Hannane et al \cite{hannane2020divide} learned a FLM topological model that performs divide-conquer search for different patches of the face using coarse to fine CNN techniques and subsequently refines the landmarks positions by using a shallow cascaded CNN regression. Gao has developed a supervised encoder-decoder architecture \cite{gao2021facial} based on EfficientNet-B0 where the dark knowledge extracted from teacher network is used to supervise the training of a small student network and patch similarity (PS) distillation is used learn the structural information of the face. 

For a given face, generated by a face detection technique, the regression model tries to estimate the position of the landmark points accurately. Thus researchers have also dedicated their time to solve the problem of face initialization. Lv et al. has developed a deep regression network with two-stage re-initialization \cite{lv2017deep} of faces at global and local scale   to unify the different face bounding boxes obtained using different face detection methods. 
Heatmap regression methods generate coarse attention maps by developing a heatmap for each landmark position \cite{newell2016stacked}. These methods \cite{chandran2020attention} \cite{dong2019teacher} \cite{kumar2020luvli} preserves the spatial features and thus provides better performance. UNet, Hourglass, Encoder-Decoder \cite{wang2019adaptive} \cite{zou2019learning} \cite{liu2019semantic} are the most widely used fully convolutional networks for generating high resolution heatmaps. Stacked hourglass network \cite{yang2017stacked}\cite{deng2019joint} have a been popular architecture to generate accurate heatmaps that can also improve the localization in case of partial occlusions. Stacking of several hourglass models improves the spatial mapping of the features in subsequent hourglasses. Stacked U-Nets have been another very popular model where the features are extracted and then attention maps are generated by performing deconvolution. Stacked densely connected U-Nets have been developed in \cite{tang2018quantized} that parses the global and local features across the U-Nets. Guo et al \cite{guo2018stacked} has defined channel aggregation block (CAB) to improve the capacity of the stacked U-Net model in parsing the features across network. In \cite{gao2020coarse} landmark-guided self attention (LGSA) block has been introduced that process the output feature map one hourglass to improve the spatial structure of the landmarks and sends it to the next hourglass. Xiong proposed a Gaussian vector \cite{xiong2020gaussian} to encode landmark coordinates that provides a better convergence result. As the heatmap-based methods requires a post-processing step which is non-differentiable, Jin has defined a single-stage pixel-in-pixel regression \cite{jin2021pixel} where for each heatmap a grid is detected on heatmap and then offsets are determined for precise localization of the landmarks.

\subsection{Attention models }
Prediction of the face landmark coordinates, using the regression techniques, generates a model that consider the locations of the landmarks. Attention methods have recently gained popularity due to the better performance in localization. The attention methods focuses on the most salient part of the face and thus reduces the unnecessary complexity due to the irrelevant parts of the image. 
Different attention models have been developed that define attention block that learns feature channel attention across the architecture, viz., squeeze-and-excitation block in \cite{hu2018squeeze}. Learning the spatial attention is very useful for facial landmark localization due to the structure of the face. Spatial attention block has been designed using non-local block in Grad-CAM \cite{selvaraju2017grad} \cite{li2019spatial}. Woo \cite{woo2018cbam} has designed CBAM that combines both spatial and channel attention. Wang  has used non-local neural net \cite{wang2018non} to learn self-attention that is most suitable due to the spatial geometry of face.
Researchers have also used adversarial training methods \cite{pan2022regressive} by introducing a discriminator module \cite{zhu2021improving} to improve the performance by generating more accurate heatmaps, followed by a attention module that optimizes the spatial correlations \cite{feng2022cascaded} between the facial landmarks. 

\section{Methodology}
We aim to develop a gaze controlled human machine interface for controlling the navigation of robotic platform and thus the primary objective is to localize the eyes of user. Thus instead of detecting 68 landmark points, we have devised our problem to detect the landmarks present only in the eye region of the face. We have selected 12 landmark points out of the 68 point landmark annotations, with indices 37 to 48. 

Attention-driven facial landmark detection is usually performed by generating ground truth heatmap. We have generated 12 groundtruth heatmaps, each for a landmark position, corresponding to 6 landmarks for each eye, by applying a gaussian filter centered at each landmark position. Standard deviation of each gaussian filter is set at 5. We have proposed an attention based model for localization of the eye heatmaps and detection of the landmarks in the eye region. The architecture of the model has been discussed in the following section.

\subsection{Network Architecture}
\begin{figure*}
	\centering
	\includegraphics[width=1.0\linewidth]{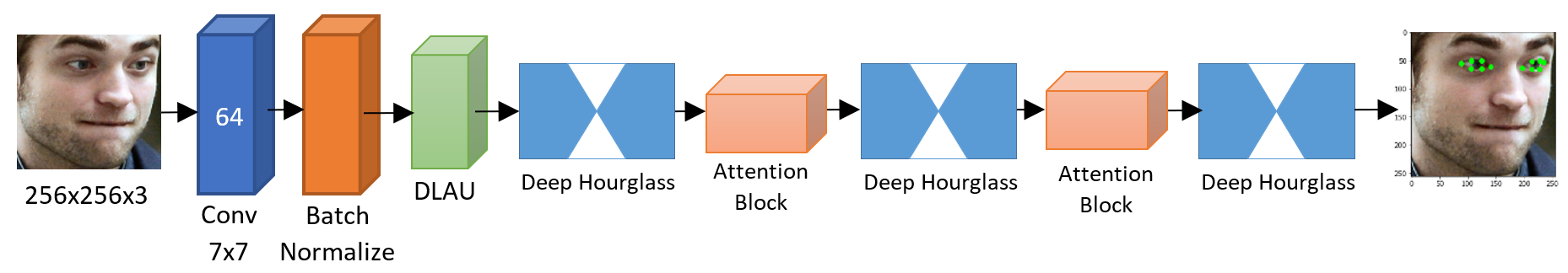}
	\caption{Framework of LocalEyenet model}
	\label{fig:networkarchitecture}
\end{figure*}

We have proposed a deep attention model called \textit{LocalEyenet} by using Stacked hourglass(HG) architecture as backbone for detection of the eye landmark positions. We have stacked 3 hourglass modules in a linear fashion as shown in Fig.\ref{fig:networkarchitecture}.
The hourglass architecture has been designed in such as way that  it combines the coarse and fine features in an efficient manner over the depth of the model. In standard hourglass architecture, the lower layer features are combined with the higher layer features using residual blocks and upsampling. The residual blocks use a skip connection to pass the low level features to the deeper layers. This alleviates the problem of vanishing gradients and also retains features even after few layers of downsampling. But this also provides a shallow aggregation between the layers and does not retain overall information across the network.

\subsubsection{Deep Hourglass architecture}
\begin{figure}
	\centering
	\includegraphics[width=0.75\linewidth]{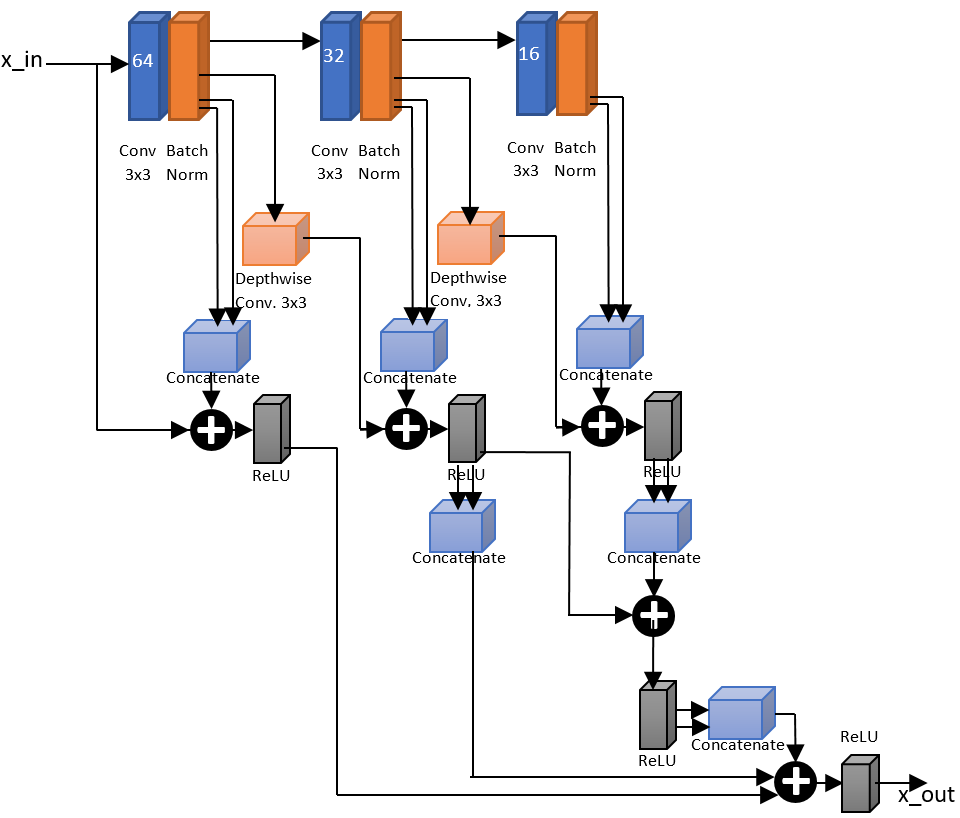}
	\caption{Deep layer aggregation unit (DLAU)}
	\label{fig:dlau}
\end{figure}

The concept of layer aggregation has been incorporated by designing deep layer aggregation schematics such as Iterative Deep Aggregation (IDA) as proposed in \cite{yu2018deep}. We have adopted the concept and designed a Deep layer aggregation unit (DLAU) that maps the features in an iterative manner. Merging the features from all modules and channels makes the feature combination more appropriate and the loss of attention over depth is being reduced. The structure of the DLAU has been described in Fig. \ref{fig:dlau}.  

\begin{figure*}
	\centering
	\includegraphics[width=1.0\linewidth]{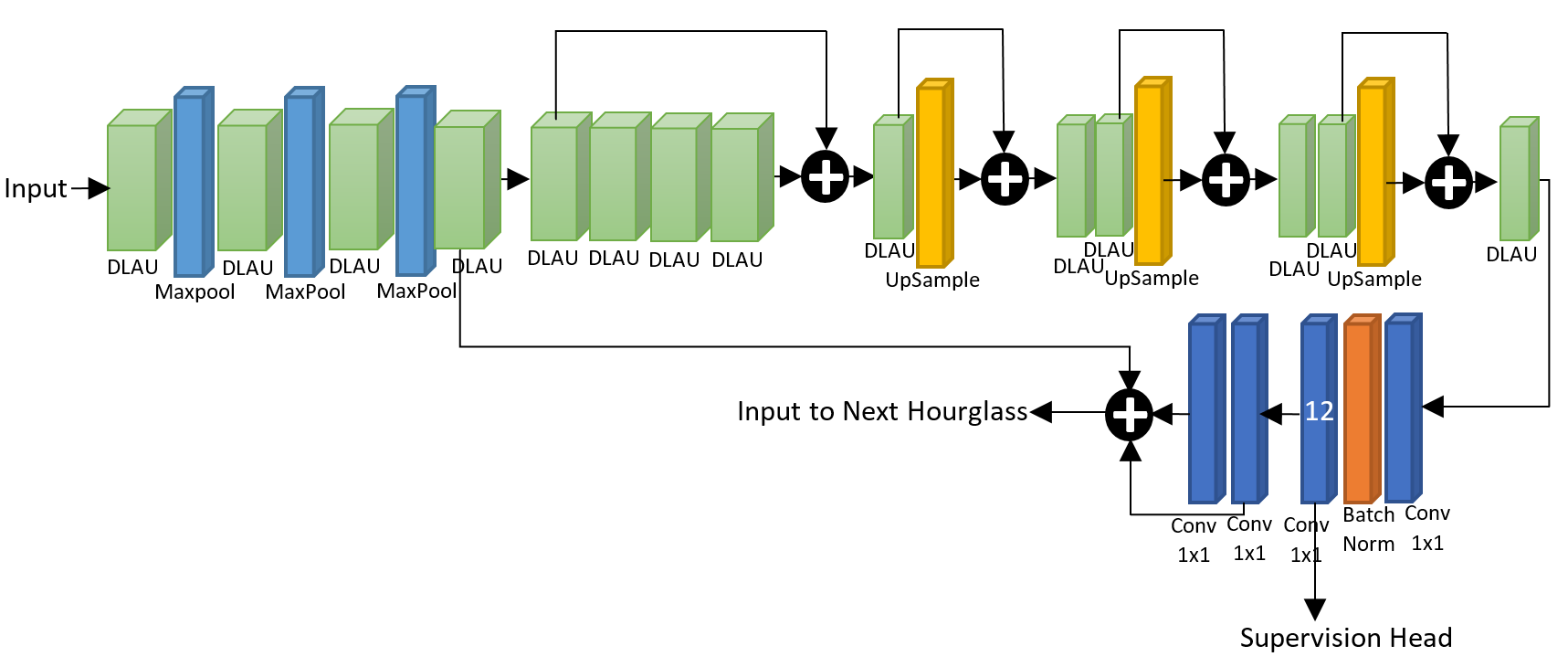}
	\caption{Architecture of Deep hourglass module with deep layer aggregation}
	\label{fig:deephourglass}
\end{figure*}

For designing our model, we have replaced the residual blocks in the hourglass modules by DLAU. Depth-wise convolution of the deep layer features are concatenated with the shallow features. This provides a deep aggregation between the features and the correlation between features across the network is preserved in this manner. The architecture of each deep hourglass module has been shown in Fig. \ref{fig:deephourglass}.

The heatmaps generated by the LocalEyenet has been passed to the post-processing stage where the positions of the landmarks are estimated from the heatmaps. Argmax is the most commonly used mathematical tool that computes the positions of the landmarks by computing the locations with the highest probability in the attention heatmap. Since argmax is non-differentiable, we have used the softargmax \cite{luvizon20182d} method which is a continuous function at each point and is thus differentiable. This allows us to train the network end to end and estimates the probability of the landmark positions on the attention heatmap in a single go.

\subsubsection{Attention between Hourglass modules}
The attention in feature maps, generated by intermediate hourglass modules, are computed and incorporated with the coarse predictions at each stage. This helps in understanding the spatial dependencies and correlation in images. We have adopted the concept of non-local neural network and modified it to compute the self-attention at each location of the feature map. 

The feature map $F_i$ obtained from the deep hourglass module is passed through a residual block to let the spatial dependencies flow across layers. The updated feature map ${F_i}^r$ is passed through a convolution layer with kernel size $1\times1$ and $12$ filters, each for a landmark point, to obtain feature map ${F_i}^{rc}$. The convolved feature map ${F_i}^{rc}$ is then passed though the softargmax operator to generate a coarse prediction of the attention map ${Map_i}$, as defined in Equation [\ref{softargmax}], where $W\times H$ is the size of the attention map. The similarity between the initial prediction of attention heatmap and the residual feature map ${F_i}^r$ is computed to rectify the attention heatmap by incorporating the spatial association by computing the element-wise dot product, as defined in Equation [\ref{ip_local}].
\begin{equation}\label{softargmax}
	Map_i =\sum_{x=1}^{W}\sum_{y=1}^{H}\frac{x}{W}\frac{y}{H}\frac{exp({F_i}^{rc}(x,y))}{\sum_{k=1}^{W}\sum_{l=1}^{H}exp({F_i}^{rc}(k,l))}
\end{equation}
\begin{equation}\label{ip_local}
	{Map_i}^1=Map_i\odot {F_i}^r
\end{equation}

\begin{figure*}
	\centering
	\includegraphics[width=1.0\linewidth]{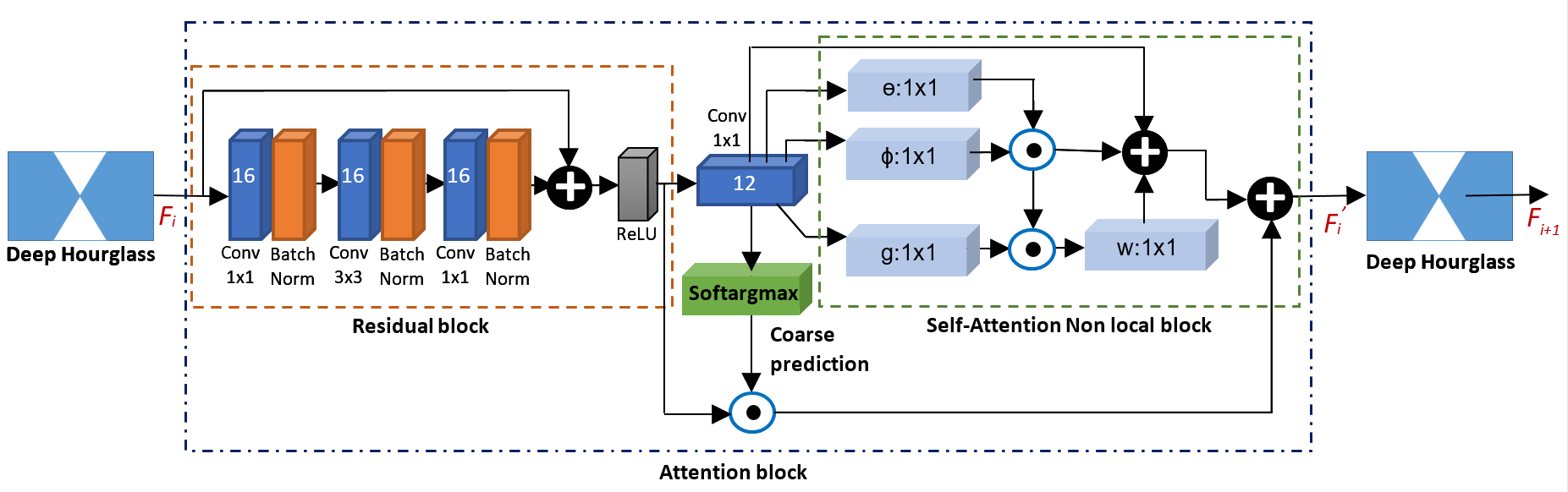}
	\caption{Design of Attention module}
	\label{fig:attentionblock}
\end{figure*}

The convolved feature map ${F_i}^{rc}$ is passed through the self-attention module. The non-local neural network has been instantiated by defining embedded Gaussian at each pixel of the image in form $\phi(x_i)=W_{\phi}x_i$, $\theta(x_j)=W_{\theta}x_j$ and $g(x_i)=W_{g}x_i$. The parameters $W_{\phi}$, $W_{\theta}$ and $W_{g}$ are learned through backpropagation. The spatial similarity between the embedded gaussians at each location is estimated by Equation [\ref{non-local1}] . The similarity between the similarity map and the convolved feature map is computed by Equation [\ref{non-local2}].
\begin{equation}\label{non-local1}
	f(i,j)= e^{\phi(x_i) \theta(x_j)^T}
\end{equation}
\begin{equation}\label{non-local2}
	S_i=softargmax(f)\odot g
\end{equation}
The non-local operation is completed by applying a transform $W$ on the self-attention map $S_i$. The residual connection ${F_i}^{rc}$ is added to the non-local output to generate the local attention map ${S_i}^{'}$ as defined in Equation [\ref{self-attention}]. The updated feature map ${F_i}^{'}$ is computed by adding the coarse prediction of the attention map ${Map_i}$, which incorporates the global attention that has been learned in the last hourglass module, to the attention map ${Att}_i$ as defined in Equation [\ref{feature_map}]. The architecture of the complete attention block is shown in Fig. \ref{fig:attentionblock}.
\begin{equation}\label{self-attention}
	{Att}_i=WS_i+ {F_i}^{rc}
\end{equation}
\begin{equation}\label{feature_map}
	{F_i}^{'} ={Att_i} +{Map_i}^1
\end{equation}

\subsubsection{Loss function}
To compute the performance of the deep modules, we have used softargmax operation to determine the coordinates $(x,y)$ of facial landmarks from the attention heatmaps. The location with the highest value of softargmax are the most probable estimates of landmark positions.

The deviation of the estimated landmark positions from the groundtruth locations is used to optimize the parameters of the model. Developed attention model aims to find local minima in the loss plot. Different loss function influences the optimized value of parameters and are computed using RMSprop optimization. We have used 3 different type of loss functions and compared the performance of the models.

As the problem is devised as a regression problem, we have first optimized the mean square error (MSE) loss as defined in Equation [\ref{mse_loss}], where L=number of landmark points. The MSE loss is computed as normalized squared L2 norm of the error function $d$ as defined in Equation [\ref{error}], where $gt$ is the groundtruth vector and $pr$ is the predicted values of the landmark points.
\begin{equation}\label{error}
	d=gt-pr
\end{equation}
\begin{equation}\label{mse_loss}
	MSE=\frac{1}{L}\sum_{i=1}^{L}{d_i}^2
\end{equation}
Presence of outliers increases the error and impairs the performance of the model. We have tried to minimize the risk of training the outliers, by learning the optimization of Huber loss, defined by Equation [\ref{huber_loss}] that computes the mean absolute error (MAE) for errors with significantly higher values.
\begin{equation} \label{huber_loss}
	L_\delta(d)=
	\begin{cases}
		\frac{d^2}{2}, & \text{if}\ \lvert d\rvert\le\delta \\
		\delta\left(\lvert d\rvert-\frac{\delta}{2}\right), & \text{otherwise}
	\end{cases}
\end{equation}
Recently, researchers have used \cite{feng2018wing}\cite{feng2020rectified} a loss function called wing loss that works well with the facial landmark detection. This loss compensates for large as well as small errors by defining a piece-wise linear and nonlinear loss function as defined in Equation [\ref{wing_loss}], where the constant $C$ smoothly connects the nonlinear and linear function.
\begin{equation}\label{wing_loss}
	Wing\left(d\right)=
	\begin{cases}
		w\ln{\left(1+\frac{ \lvert d\rvert}{\epsilon}\right)}, & \text{if}\  \lvert d\rvert\le\delta \\
		\left( \lvert d\rvert-C\right)\ , & \text{otherwise}
	\end{cases}
\end{equation}
\begin{equation}\label{C}
	C=w-wln(1+\frac{w}{\epsilon})
\end{equation}

The developed coarse-to-fine architectures models have been optimized using different loss functions and the performance have been discussed in the subsequent results section.

\section{Results and Discussion}
The faces are first detected and cropped for detection of facial landmarks. High resolution face images are downsampled and smaller images are upsampled to the size $256\times256$. The resized images are passed through developed deep learning framework.

\subsection{Dataset}
We have used 2 public domain datasets in this paper, 300W and 300VW. The results of the designed attention model architecture has been trained on the images from the 300W dataset and has been tested on both the datasets.
\subsubsection{300W}
The 300W dataset \cite{Sagonas_2013_CVPR_Workshops} \cite{Sagonas_2013_ICCV_Workshops} \cite{sagonas2016300} contains facial images captured in two environmental settings, namely indoor and outdoor conditions. This ensures the varied illumination conditions, large variation in expression, pose and occlusion that are being covered in the dataset formation. The dataset contains total 600 images with 300 in indoor and 300 in outdoor categories. Faces in the images are annotated using 68 landmark points using a semi-automatic methodology. The images with multiple number of faces have been annotated separately with different annotation files. 

\subsubsection{300VW}
The 300VW dataset \cite{chrysos2015offline}\cite{shen2015first}\cite{tzimiropoulos2015project} contains facial videos of 113 subjects recorded in wild scenarios at varied frame rate of 25-30 fps, where each video is around 1 minute long. All the frames have been annotated using 68 landmark points. The videos are recorded under three different scenarios such as well-lit conditions, different illumination conditions and completely unconstrained conditions including occlusions, make-up, expression, head pose, etc. We have extracted frames from the videos and cropped the faces from each frame. The faces has been reshaped later and the annotations have been updated accordingly as discussed in the data pre-processing section.

\subsection{Data Preprocessing}
The images are of different resolution and the face contains a small part of the image in most of the cases. Some of the images contain more than one faces and thus we cannot use the whole image for designing the architecture. Thus faces are cropped from the images and resized to the resolution $256\times256$. The landmark annotations are also scaled accordingly as defined in Equation [\ref{landmarks_rescaling}], where $(x,y)$ are the coordinates of any landmark point and $(x_{new},y_{new})$ are the coordinates of newly mapped location of the landmark and  $[h,w]$ and $[h_{new},w_{new}]$ are the height and width of the original image and resized image respectively.
\begin{equation} \label{landmarks_rescaling}
	\begin{aligned}
		x_{new} &= x*(w_{new}/w)\\
		y_{new} &= y*(h_{new}/h)
	\end{aligned}
\end{equation}

We aim to localize the eyes of the user for further gaze tracking operation. Thus we have selected only 12 landmark points that represent both the eyes on the faces. We initialize heatmaps for each landmark position by defining a Gaussian centered at each landmark with standard deviation of 5. Thus for each input image we generate 12 heatmaps, each centered at one of the 12 landmark points.

\subsection{Data Augmentation}
We have augmented the images by doing different mathematical operations such as horizontal flipping and rotation. We have also performed blurring on the face images to make a robust dataset for training. 
The landmark points of the horizontally flipped images are redefined in Equation [\ref{landmarks_hf}].
\begin{equation} \label{landmarks_hf}
	\begin{aligned}
		x_{new} &= w_{new}-x\\
		y_{new} &= y
	\end{aligned}
\end{equation}
The cropped and resized faces have also been rotated by very small angle and the rotated faces have been added to the augmented dataset. The rotation angle has been selected as $-5^\circ$, $+5^\circ$, $-10^\circ$ and $+10^\circ$ and a rotated image has been generated in each category. The landmark points have been rescaled to the new coordinate system as defined in Equation [\ref{landmarks_rotate}]. 
\begin{equation} \label{landmarks_rotate}
	\begin{aligned}
		x_{new} &= x \cos{\theta}+y \sin{\theta} +x_{offset}\\
		y_{new} &= -x \sin{\theta}+y \cos{\theta} +y_{offset}
	\end{aligned}
\end{equation}
We have used Gaussian filter of dimension $9\times9$ with standard deviation $\sigma_x=\sigma_y=1.8$ for generating blurred images. The filter is defined as Equation [\ref{gauss_filter}].
\begin{equation} \label{gauss_filter}
	f(x,y)=\frac{1}{\sqrt{2\pi\sigma_x\sigma_y}}\exp^{-(x^2+y^2)/{2\sigma_x\sigma_y}}
\end{equation}

There are few images, generated after rotation, which are partially cropped, and this leads to the cropping of some landmark points. Thus these images have been manually checked and removed from the dataset. The generated augmented 300W dataset contains 4195 images in total, where the distribution of the images in each category is defined in Table [\ref{data_aug}].
\begin{table}[h!]
	\begin{center}
		\caption{Images generated after data augmentation of 300W dataset}
		\label{data_aug}
		\begin{tabular}{l|c|c|c} 
			\hline
			{Category} & {Indoor} & {Outdoor} & {Total}\\
			\hline
			Original & 300 & 300 & 600\\
			Horizontally Flipped & 1197 & 1198 & 2395\\
			Rotated  & 300 & 300 & 600\\
			Blurred  & 300 & 300 & 600\\
			\hline
		\end{tabular}
	\end{center}
\end{table}

\subsection{Evaluation}
\begin{figure*}[!t]
	\centering
	\subfloat[]{\includegraphics[width=0.6in]{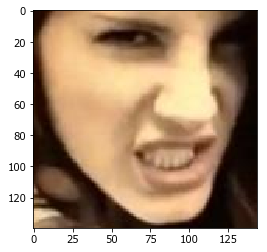}}
	\hfil
	\subfloat[]{\includegraphics[width=0.59in]{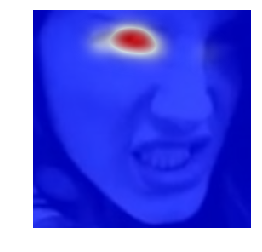}}
	\hfil
	\subfloat[]{\includegraphics[width=0.59in]{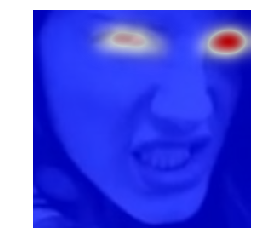}}
	\hfil
	\subfloat[]{\includegraphics[width=0.59in]{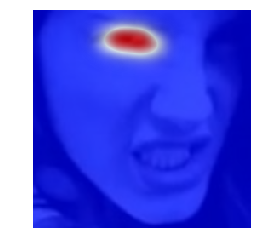}}
	\hfil
	\subfloat[]{\includegraphics[width=0.59in]{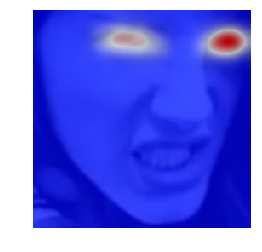}}
	\hfil
	\subfloat[]{\includegraphics[width=0.59in]{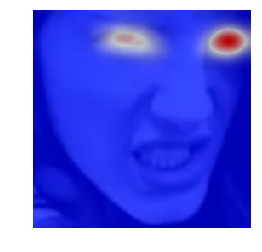}}
	\hfil
	\subfloat[]{\includegraphics[width=0.59in]{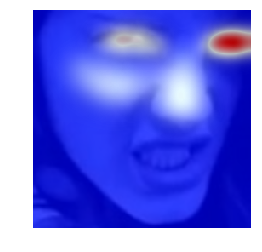}}
	\hfil
	\subfloat[]{\includegraphics[width=0.59in]{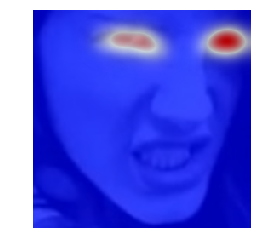}}
	\hfil
	\subfloat[]{\includegraphics[width=0.59in]{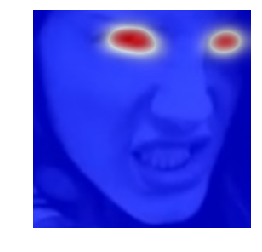}}
	\vfil
	\subfloat[]{\includegraphics[width=0.6in]{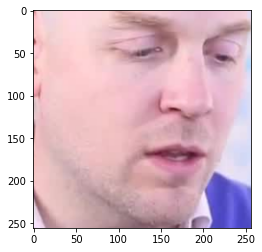}}
	\hfil
	\subfloat[]{\includegraphics[width=0.59in]{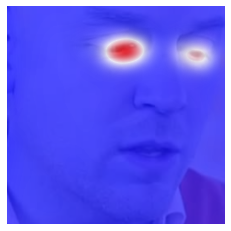}}
	\hfil
	\subfloat[]{\includegraphics[width=0.59in]{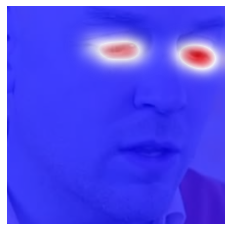}}
	\hfil
	\subfloat[]{\includegraphics[width=0.59in]{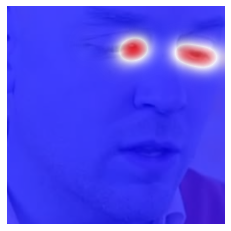}}
	\hfil
	\subfloat[]{\includegraphics[width=0.59in]{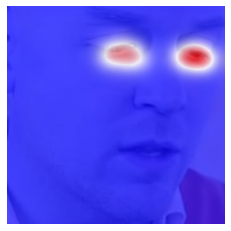}}
	\hfil
	\subfloat[]{\includegraphics[width=0.59in]{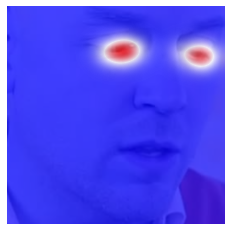}}
	\hfil
	\subfloat[]{\includegraphics[width=0.59in]{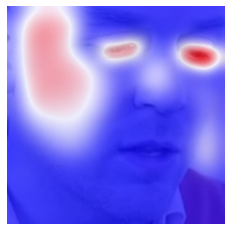}}
	\hfil
	\subfloat[]{\includegraphics[width=0.59in]{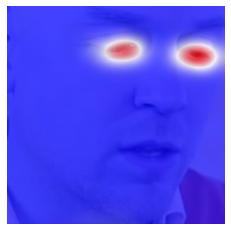}}
	\hfil
	\subfloat[]{\includegraphics[width=0.59in]{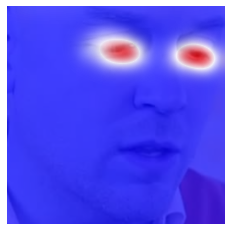}}
	
	\caption{Attention Heatmaps generated by different deep learning frameworks on sample image from 300W and 300VW dataset. (a-i) corresponds to image from 300W dataset and (j-r) corresponds to image from 300VW dataset; (a),(j) Original image, (b),(k) Hourglass model, (c),(l) Unet model, (d),(m) DenseUnet model, (e),(n) Stacked Unet model, (f),(o) Stacked Hourglass model, (g),(p) Densely connected Unet model \cite{tang2018quantized}, (h),(q) Stacked Hourglass with CAB\cite{guo2018stacked}, (i),(r) LocalEyenet Model}
	\label{fig:heatmaps_generated}	
\end{figure*}

\begin{figure*}[!t]
	\centering
	\subfloat[]{\includegraphics[width=0.45\linewidth,height=2in]{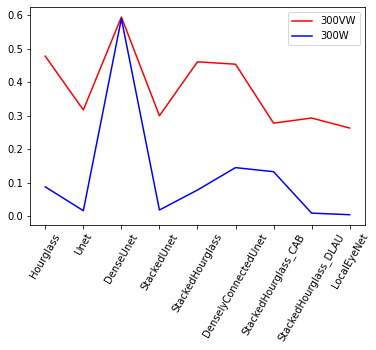}}
	\hfil
	\subfloat[]{\includegraphics[width=0.5\linewidth,height=2in]{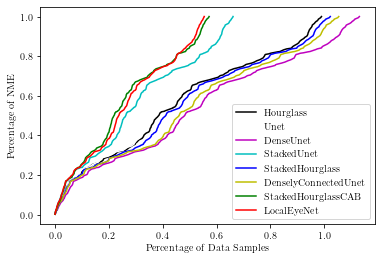}}
	\caption{a. NME Plot, b.Cumulative Error Distribution curves for landmark localization on 300W dataset} 
	\label{fig:NMEPlot}	
\end{figure*}

We have developed a coarse-to-fine architecture incorporating attention and have evaluated the performance of the architecture in terms of Normalized Mean Error(NME) and Area under Curve(AUC). NME is defined in Equation [\ref{nme}] where $iod$ is the inter-ocular distance that is defined as L2 norm between the outer corners of  eyes. 
\begin{equation} \label{nme}
	NME=\frac{1}{L}\sum_{i=1}^{L}\frac{{\|gt_i-pr_i\|} _2}{iod}
\end{equation}
The AUC and failure rate (FR) is computed from the cumulative error distribution curve with threshold NME set at 0.05. The performance of the different architectures have been enlisted in Table [\ref{performance}]. The attention heatmaps generated by different frameworks on sample images of 300W and 300VW have been shown in Fig. \ref{fig:heatmaps_generated}. The NME obtained for 300W and 300VW datasets evaluated using different models have been shown in Fig. \ref{fig:NMEPlot}.
\begin{table}
	\begin{center}
		\caption{Performance of different deep learning framework on 300W and 300VW datasets}
		\label{performance}
		\begin{tabular}{l | c|c|c}
			\hline
			{\textbf{Methodology}} & \multicolumn{2}{c}{\textbf{300W}} & \multicolumn{1}{|c}{\textbf{300VW}}\\
			\cline{2-4}
			&  {\textbf{NME}} & {AUC} & {\textbf{NME}}\\
			\hline
			Hourglass model & 0.0878 & 0.1672 & 0.4781\\
			Unet model & 0.0167 & 0.8825  & 0.3176\\
			DenseUnet model  & 0.5901 & 0  & 0.5950\\
			Stacked Unet model  & 0.0187 & 0.6371  & 0.3002\\
			Stacked Hourglass model & 0.0784 & 0.1948  & 0.4612\\
			Densely Connected Unet\cite{tang2018quantized} & 0.1453 & 0.0611 & 0.4541\\
			Stacked HG with CAB\cite{guo2018stacked} & 0.1331 & 0.0761  & 0.2781\\
			Stacked Hourglass with DLAU  & 0.0094 & 0.8145 & 0.2934\\
			\textbf{LocalEyenet model} & \textbf{0.0047} & \textbf{0.9082} &  \textbf{0.2635}\\
			\hline
		\end{tabular}
	\end{center}
\end{table}

%
%
%

Three different losses, viz., MSE, huber and wing loss have been optimized for determination of optimum parameters for the developed deep learning framework. The performance of the models for 3 loss functions has been enlisted in Table [\ref{loss_comp}]. As we can visualize from the table, for most of the frameworks, including our architecture, the NME is minimum for MSE loss optimization. Huber loss also provides good generalization for Hourglass model, DenseUnet model and  Stacked HG model with CAB\cite{guo2018stacked}. NME obtained for wing loss optimization are comparatively quite high for almost all the models. Thus we have selected MSE loss for optimization of the model parameters. 
\begin{sidewaystable}
	\caption{Evaluation of NME for different loss functions on 300W dataset}
	\label{loss_comp}
	\begin{center}
		\begin{tabular}{l|c|c|c|c|c|c|c|c|c}
			\hline
			{\textbf{Methodology}} & \multicolumn{3}{c}{\textbf{MSE loss}} & \multicolumn{3}{|c}{\textbf{Huber loss}} & \multicolumn{3}{|c}{\textbf{Wing loss}}\\
			\cline{2-10}
			&  {NME} & {AUC} & {FR} & {NME} & {AUC} & {FR} & {NME} & {AUC} & {FR}\\
			\hline
			Hourglass model & 0.0878 & 0.1672 & 0.5983 & \textbf{0.0577}	& 0.2786 & 0.4233 & 8.8091 & 0 & 1\\
			Unet model & \textbf{0.0167} & 0.8825 & 0.0167 & 0.3047 & 0.0015 & 0.9817 & 9.7909 & 0 & 1\\
			DenseUnet model  & 0.5901 & 0 & 1 & \textbf{0.5120} & 0 & 1 & 9.7909 & 0 & 1\\
			Stacked Unet model  & \textbf{0.0187} & 0.6371 &  0.0467 &  0.0187 & 0.6370 &  0.0469 & 9.7909 & 0 & 1\\
			Stacked Hourglass model & \textbf{0.0784} & 0.1948 & 0.5533 & 0.3349	& 0.0004 & 0.9916 & 4.6254& 0& 1\\
			Densely Connected Unet model\cite{tang2018quantized} & \textbf{0.1453} & 0.0611 & 0.7750 & 3.761 & 0 & 1 & 9.7909 & 0 &	1\\
			Stacked Hourglass with CAB\cite{guo2018stacked} & 0.1331 & 0.0761 & 0.7417 & \textbf{0.0777} & 0.1972 &	0.5533 & 0.3232 & 0.0007 & 0.99\\
			\textbf{LocalEyenet model} & \textbf{0.0047} & 0.9082 & 0 & 0.0915 & 0.1573 & 0.6050 & 0.1516 & 0.0542 &  0.7883\\
			\hline
		\end{tabular}
	\end{center}
\end{sidewaystable}

The cumulative mean error is computed over all the samples of 300W dataset. The Cumulative Error Distribution (CED) curve as shown in Fig. \ref{fig:NMEPlot}, shows that the DenseUnet model generates the highest NME and our LocalEyenet model displays the lowest NME and covers the maximum AUC.

\section{Ablation study}
In order to understand the role of attention block in the eye localization architecture, we have analyzed the model performance for different scenarios of 300VW dataset. We have implemented standard state of the art Stacked Hourglass model and introduced the DLAU unit in place of Residual block in each hourglass. The performance of model is compared by computing the NME, failure rate and AUC for the models. Introduction of the attention block between the Deep hourglass modules improves the localization of the heatmaps in the annotated area. Comparison of the performance for ablation study has been enlisted in Table [\ref{ablation_300W}] for 300W dataset.
\begin{table*}
	\caption{Ablation study: Evaluation of NME on 300W dataset}
	\label{ablation_300W}
	\begin{center}
		\begin{tabular}{l|c|c|c|c|c|c}
			\hline
			{\textbf{Methodology}} & \multicolumn{2}{c}{\textbf{MSE loss}} & \multicolumn{2}{|c}{\textbf{Huber loss}} & \multicolumn{2}{|c}{\textbf{Wing loss}}\\
			\cline{2-7}
			&  {NME} & {AUC} & {NME} & {AUC} & {NME} & {AUC}\\
			\hline
			Stacked Hourglass model & \textbf{0.0784} & 0.1948 & 0.3349	& 0.0004 & 4.6254& 0\\
			Stacked Hourglass with DLAU  &  \textbf{0.0094} & 0.8145 & 0.0095 & 0.8143 &9.7909	&0\\
			\textbf{LocalEyenet model} & \textbf{0.0047} & 0.9082 & 0.0915 & 0.1573 & 0.1516 & 0.0542\\
			\hline
		\end{tabular}
	\end{center}
\end{table*}

\begin{figure*}[!t]
\centering
\subfloat[]{\includegraphics[width=1.0in]{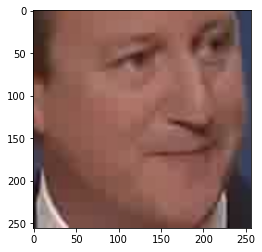}}
\hfil
\subfloat[]{\includegraphics[width=1.0in]{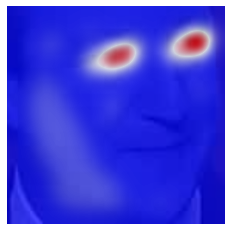}}
\hfil
\subfloat[]{\includegraphics[width=1.0in]{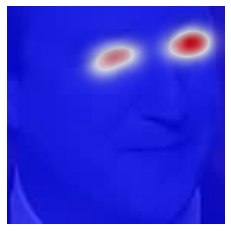}}
\hfil
\subfloat[]{\includegraphics[width=1.0in]{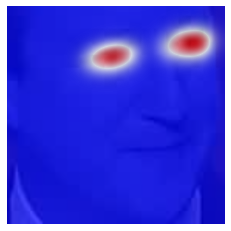}}
\vfil
\subfloat[]{\includegraphics[width=1.0in]{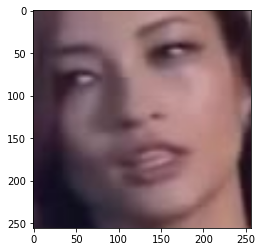}}
\hfil
\subfloat[]{\includegraphics[width=1in]{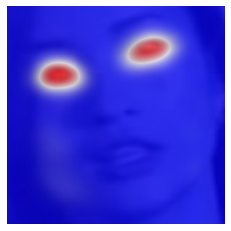}}
\hfil
\subfloat[]{\includegraphics[width=1in]{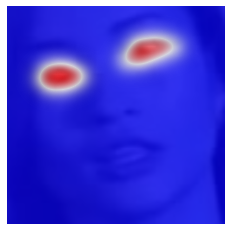}}
\hfil
\subfloat[]{\includegraphics[width=1in]{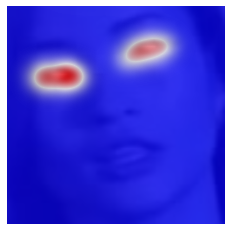}}
\vfil
\subfloat[]{\includegraphics[width=1in]{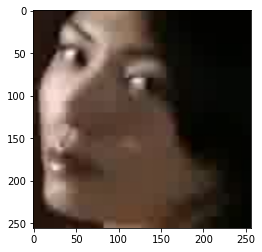}}
\hfil
\subfloat[]{\includegraphics[width=1in]{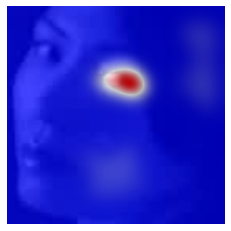}}
\hfil
\subfloat[]{\includegraphics[width=1in]{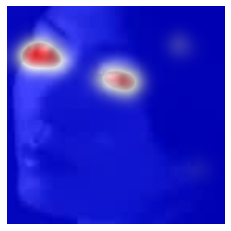}}
\hfil
\subfloat[]{\includegraphics[width=1in]{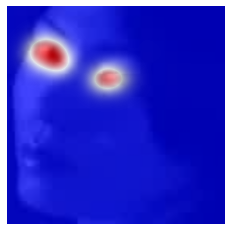}}
\caption{Attention Heatmaps for ablation study on sample images from 3 scenarios of 300W dataset. (a) Image from Scenario 1, (e) Image from Scenario 2, (i) Image from Scenario 3, (b),(f),(j) Heatmap generated by Stacked Hourglass model on image from Scenario 1, Scenario 2, Scenario 3 respectively, (c),(g),(k) Heatmap generated by Stacked Hourglass with DLAU on image from Scenario 1, Scenario 2, Scenario 3 respectively, (d),(h),(l) Heatmap generated by LocalEyenet Model on image from Scenario 1, Scenario 2, Scenario 3 respectively.}
\label{fig:ablation_300VW_scenarios}	
\end{figure*}

The videos in 300VW dataset contains three different scenarios. Videos from scenario 1 have almost no illumination variation, but contain head pose variations. Videos from scenario 2 have been recorded in unconstrained conditions with different illuminations and videos from scenario 3 have all variations including illumination, occlusions, make-up and head pose. The performance of the attention model and its ablation for the 3 different scenarios have been assessed in Table [\ref{ablation_300VW}]. The attention heatmaps generated by standard stacked hourglass model, the updated stacked hourglass model with DLAU and the LocalEyenet model for the 3 different scenarios have been shown in Fig. \ref{fig:ablation_300VW_scenarios}. 
\begin{table}
	\caption{Evaluation of NME for different scenarios on 300VW dataset}
	\label{ablation_300VW}
	\begin{center}
		\begin{tabular}{l|c|c|c}
		\hline
		{\textbf{Methodology}} & \multicolumn{3}{c}{\textbf{NME}}\\
		\cline{2-4}
	&  {Scenario1} & {Scenario2} & {Scenario3} \\
	\hline
	Stacked Hourglass & 0.0499 & 0.0594 & 0.3531\\
	Stacked HG with DLAU  & 0.0446 & 0.0589 & 0.0443 \\
	\textbf{LocalEyenet model} & \textbf{0.0283} & \textbf{0.0371} & \textbf{0.0402} \\
	\hline
\end{tabular}
\end{center}
\end{table}

\section{Testing on real-time video stream}
\begin{figure*}[!t]
\centering
\subfloat[]{\includegraphics[width=1in]{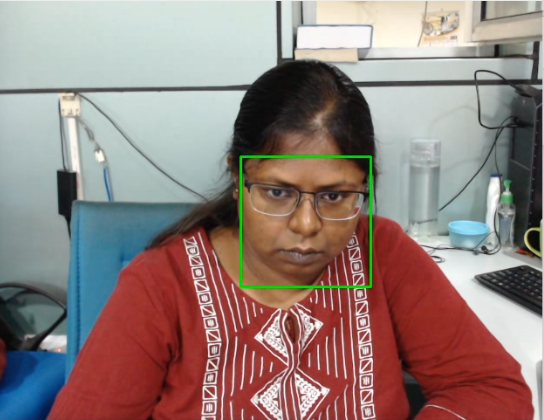}}
\hfil
\subfloat[]{\includegraphics[width=1in]{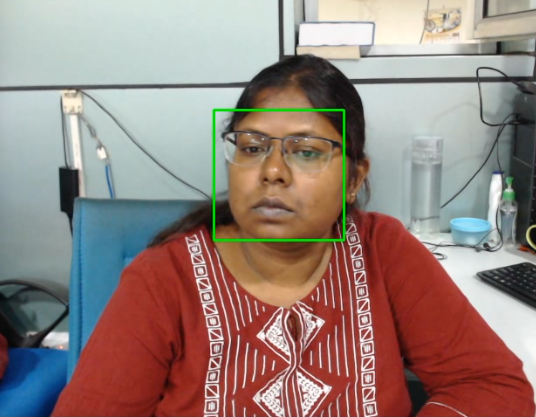}}
\hfil
\subfloat[]{\includegraphics[width=1in]{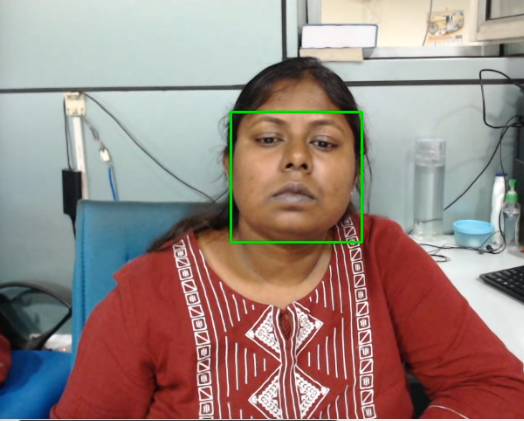}}
\hfil
\subfloat[]{\includegraphics[width=1in]{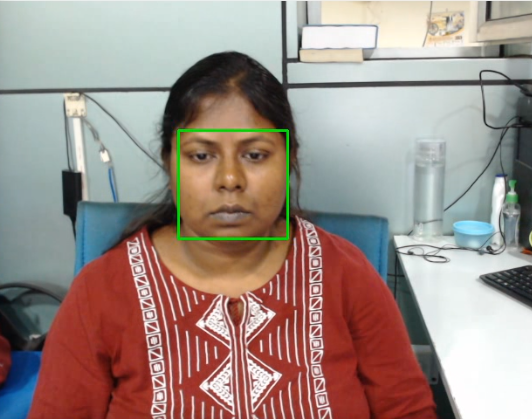}}
\hfil
\subfloat[]{\includegraphics[width=1in]{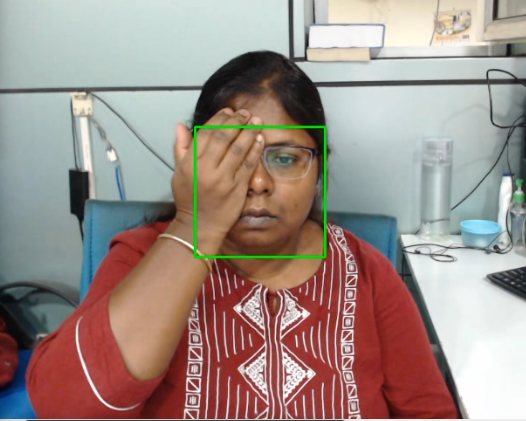}}
\hfil
\subfloat[]{\includegraphics[width=1in]{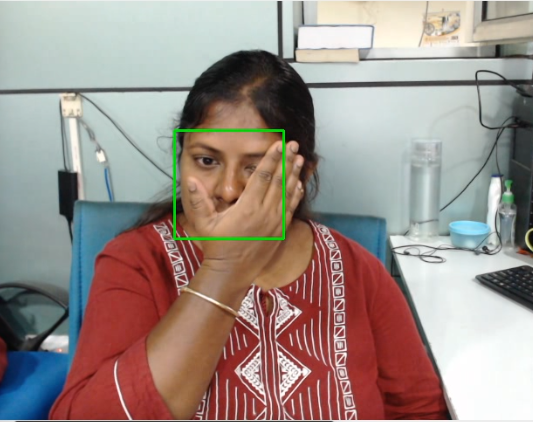}}
\hfil
\subfloat[]{\includegraphics[width=0.8in]{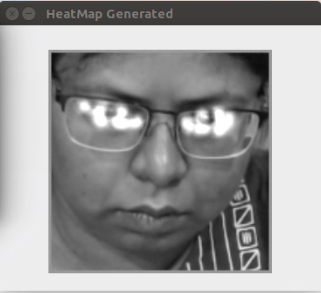}}
\hfil
\subfloat[]{\includegraphics[width=0.8in]{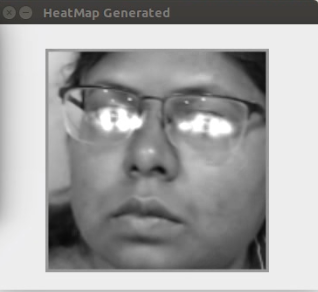}}
\hfil
\subfloat[]{\includegraphics[width=0.8in]{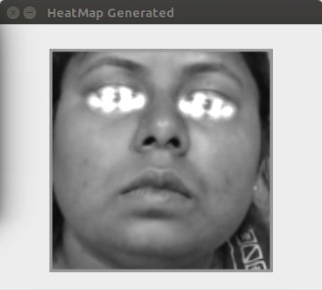}}
\hfil
\subfloat[]{\includegraphics[width=0.8in]{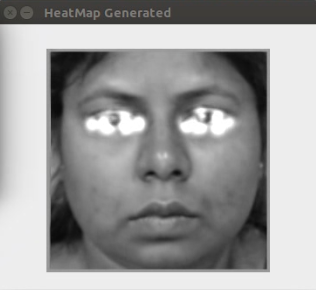}}
\hfil
\subfloat[]{\includegraphics[width=0.8in]{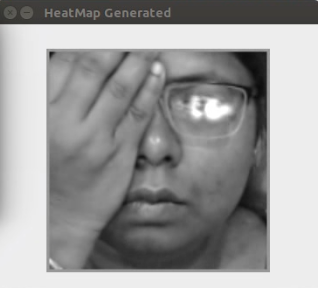}}
\hfil
\subfloat[]{\includegraphics[width=0.8in]{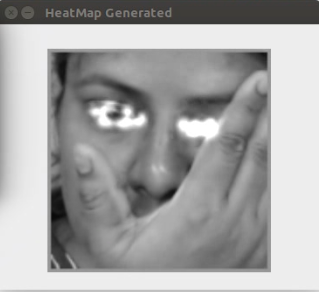}}
\caption{Detection of eye landmarks in frames of real-time video. (a)-(f) frame1 to frame6, (g)-(l) Heatmaps generated by LocalEyenet Model for frame1 to frame6.}
\label{fig:realtime_Testing}	
\end{figure*}

We have tested the performance of our model in real-time video stream. The real-time video streaming has been performed using logitech C920 pro webcam which captures frames at 35 fps with resolution of $1920\times 1080$. The faces in each frame of the video stream is initialized using dlib face detector. The detected faces are passed through the LocalEyenet model for inferencing. We have tried to impose different conditions such as head pose variation, removal of spectacles and partial occlusion using hands etc. to check the performance of our model. The inference has been performed at the rate of 32 fps which is very close to the frame rate of the videos. The mean error corresponding to the frames are very low and the eye landmarks detected on the faces in different frames have been displayed in Fig.\ref{fig:realtime_Testing}.
\begin{figure*}[!t]
\centering
\subfloat[]{\includegraphics[width=1.5in]{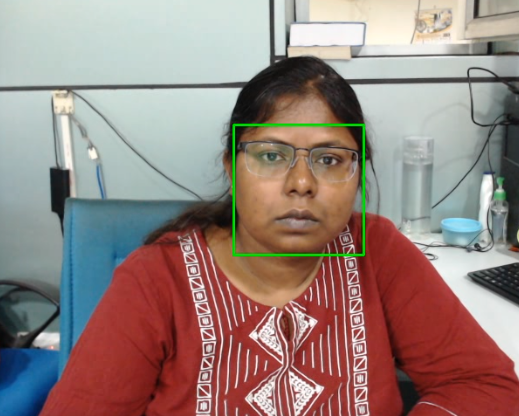}}
\hfil
\subfloat[]{\includegraphics[width=1.2in]{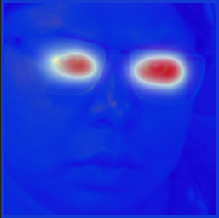}}
\caption{Eye heatmap on frame of real-time video. (a) Real-time image frame, (b) Heatmap.}
\label{realtime_Testing}
\end{figure*}

\section{Conclusion}
Gaze detection systems have the very potential to develop a human-machine interface device. To build a gaze detection system, it is of utmost necessity to develop a machine learning framework using facial landmarks. In this work, we propose a convolutional framework named LocalEyenet. The framework is an attention driven neural network architecture that focuses on localization of region of interests in image. The model detects the facial landmarks corresponding to eye regions using heatmap based regression. We show that our framework performs better than any other state of the art heatmap regression method, even with variations in illumination, pose and occlusion by spectacles and hand. Our model learns to localize the heatmaps at an early stage and thus generates inferences at a very high speed. To summarize, LocalEyenet is a robust facial landmark localizer that can perform very well in real-time eye localization under different varied environment conditions.

\section*{Acknowledgment}
The authors would like to acknowledge Council of Scientific \& Industrial Research (CSIR)-Central Electronic Engineering Research Institute (CEERI), Pilani, Rajasthan, India for providing the facilities for conducting the research work. 

\bibliographystyle{elsarticle-harv} 
\bibliography{refbank}

\end{document}